\documentclass[conference, a4paper]{IEEEtran}

\usepackage{amsmath,amsfonts,bm}

\newcommand{\eqdef}{\ensuremath{\stackrel{\mbox{\upshape\tiny def}}{=}}}

\def\1{\bm{1}}

\def\rvs{{\mathbf{s}}}

\def\rvx{{\mathbf{x}}}
\def\rvy{{\mathbf{y}}}
\def\rvz{{\mathbf{z}}}

\DeclareMathAlphabet{\mathsfit}{\encodingdefault}{\sfdefault}{m}{sl}
\SetMathAlphabet{\mathsfit}{bold}{\encodingdefault}{\sfdefault}{bx}{n}

\newcommand{\R}{\mathbb{R}}

\DeclareMathOperator*{\argmax}{arg\,max}
\DeclareMathOperator*{\argmin}{arg\,min}

\usepackage{xcolor}
\usepackage{balance}
\usepackage[pagebackref=true,breaklinks=true,colorlinks,bookmarks=false]{hyperref}
\usepackage[capitalize]{cleveref}

\ifCLASSINFOpdf
  \usepackage[pdftex]{graphicx}
\else
  \usepackage[dvips]{graphicx}
\fi
\usepackage{amsmath}
\ifCLASSOPTIONcompsoc
 \usepackage[caption=false,font=normalsize,labelfont=sf,textfont=sf]{subfig}
\else
 \usepackage[caption=false,font=footnotesize]{subfig}
\fi

\begin{document}

\title{Deep Variational Inverse Scattering}

\author{\IEEEauthorblockN{
AmirEhsan Khorashadizadeh\IEEEauthorrefmark{1},   
Ali Aghababaei\IEEEauthorrefmark{2},   
Tin Vlašić\IEEEauthorrefmark{3},
Hieu Nguyen\IEEEauthorrefmark{1},    
Ivan Dokmanić\IEEEauthorrefmark{1}      
}                                     
\IEEEauthorblockA{\IEEEauthorrefmark{1}
Department of Mathematics and Computer Science, University of Basel, Basel, Switzerland}
\IEEEauthorblockA{\IEEEauthorrefmark{2}
Department of Electrical Engineering, Sharif University of Technology}
\IEEEauthorblockA{\IEEEauthorrefmark{3}
Faculty of Electrical Engineering and Computing, University of Zagreb, Zagreb, Croatia}
\IEEEauthorblockA{ \emph{*amir.kh@unibas.ch} }
}


\maketitle

\begin{abstract}

Inverse medium scattering solvers generally reconstruct a single solution without an associated measure of uncertainty. This is true  both for the classical iterative solvers and for the emerging deep-learning methods. But ill-posedness and noise can make this single estimate inaccurate or misleading. While deep networks such as conditional normalizing flows can be used to sample posteriors in inverse problems, they often yield low-quality samples and uncertainty estimates. 
In this paper, we propose U-Flow, a Bayesian U-Net based on conditional normalizing flows, which generates high-quality posterior samples and estimates  physically-meaningful uncertainty. We show that the proposed model significantly outperforms the recent normalizing flows in terms of posterior sample quality while having comparable performance with the U-Net in point estimation. Our implementation is available at \url{https://github.com/swing-research/U-Flow}.
\end{abstract}

\vskip0.5\baselineskip

%

\section{Introduction}
In inverse medium scattering, the goal is to determine the properties of an unknown model from the measured scattered fields. The unknown model parameters is the medium containing scatters. Inverse scattering has numerous applications, including radar imaging~\cite{cheney2009fundamentals}, through-wall imaging~\cite{amin2017through}, medical ultrasound imaging, geophysics~\cite{persico2014introduction}, and nondestructive testing~\cite{zoughi2000microwave}. We consider the reconstruction of a finite number of parameters of the model from the scattered fields.
The forward model, given by a partial differential equation, is a mathematical operator mapping the model parameters  to the measurements. In our case, the forward model is the time-harmonic wave equation and is nonlinear. Even though it is a linear equation in the source term, it is nonlinear in the model parameters. The nonlinearity is due to the multiple scattering and complicates the inversion; the problem becomes more nonlinear as the contrast increases~\cite{chen2018computational}. Moreover, while inverse medium scattering is well-posed and Lipschitz stable for continuous measurements~\cite{nachman1996global}, it is an ill-posed inverse problem for sparse finite of measurements: there are many plausible reconstructions that fit the measurements to within the noise level.

This variety of solutions suggests using methods that recover more than a single reconstruction~\cite{sun2021deep, khorashadizadeh2022conditional}. A probabilistic characterization of solutions enables us more reliable interpretation of reconstructions and gives an important measure of uncertainty as the problem gets more ill-posed. Estimating uncertainty is paramount in safety-critical tasks such as medical imaging \cite{begoli2019need}, nuclear stockpile \cite{stoyer2009science,brown2015uncertainty}, and more recently self-driving vehicles \cite{arnez2020comparison,michelmore2018evaluating}.

It is natural to use a Bayesian approach via modeling the posterior over reconstructions.
We assume that the measurements $y$ and the unknown model parameters $x$ are realizations of random vectors, ${Y \in \mathcal{Y}}$, ${X \in \mathcal{X}}$ with a joint distribution $p_{X,Y}$. The posterior distribution $p_{X|Y}$ is the conditional probability of the parameters given the observed measurements and can be expressed using Bayes rule as
\begin{equation}
    p_{X|Y}(x|y) = \dfrac{p_{Y|X}(y|x) p_X(x)}{\int_x p_{X,Y}(x,y)  dx}.
\end{equation}
In real-world high-dimensional imaging problems, computing $\int_x p_{X,Y}(x,y)  dx$ is intractable. Moreover, the prior distribution of the unknown parameters $P_X$ is unknown and must be estimated.

There are a number of approaches to approximate or sample the posterior. Tarantola \cite{tarantola2005inverse}, as well as Stuart \cite{stuart2010inverse}, provided a comprehensive review of inverse problems from a statistical point of view. Traditional approaches include variants of Markov chain Monte Carlo \cite{martin2012stochastic,zhao2019gradient} which exploits the operator structure. The main challenge is a large number of required forward simulations. To alleviate the computational cost, a class of methods \cite{cui2015data,peherstorfer2018survey} employ data-driven model reduction. More recently, neural networks have shown promising results for posterior approximation. Variational U-Net is proposed by Esser \emph{et al.}~\cite{esser2018variational} to generate images from poses and exploited by Jin \emph{et al.}~\cite{jin2019fast} for reservoir simulations where the network is trained with the evidence lower bound (ELBO), similar to variational autoencoders~\cite{kingma2013auto}.
Bayesian convolutional neural networks are used for posterior sampling in several computational imaging problems~\cite{siahkoohi2022deep,wei2020uncertainty}.

Conditional normalizing flows~\cite{ardizzone2019guided} are a class of deep generative models to approximate the posterior. They provide efficient posterior sampling, likelihood estimation and uncertainty quantification~\cite{siahkoohi2020faster,zhao2022bayesian}. However, they have their own drawbacks: they require significant memory and are slow to train. Moreover, they lack any architectural regularization over posterior samples, yielding low-quality reconstructions in ill-posed inverse problems. Conditional injective flows~\cite{khorashadizadeh2022conditional} remedy these drawbacks by using a low-dimensional latent space; they can be trained fast and have a low-memory footprint leading to better posterior samples. Still, the quality of reconstructions is inferior to highly successful image-to-image regression models like the U-Net~\cite{ronneberger2015u}, especially for non-linear inverse problems.

In this paper, we propose U-Flow, a Bayesian U-Net based on conditional normalizing flows. U-Flow benefits from favorable aspects of the U-Net: it yields high-quality reconstructions even for non-linear inverse problems, while enabling regularized posterior sampling and meaningful uncertainty estimates. We show that the MMSE estimate from U-Flow has comparable quality to that of U-Net and it significantly outperforms conditional injective flows in posterior sampling and uncertainty quantification.

\section{Wave Scattering Model}

We use the Helmholtz (time-harmonic wave) equation as the forward model,\footnote{Here we numerically solve the equation using the \texttt{j-wave} package \cite{stanziola2022jwave}.}
\begin{equation}\label{eq:helmholtz_eq}
    \triangle u - \dfrac{\omega^2}{c^2} u = -i g ~\text{in}~ \Omega.
\end{equation}
The medium is characterized by the wave speed $c$. In this work, we restrict ourselves to the structural heterogeneity in the wave speed, while leaving the density constant and the attenuation zero. The source $g$ is a point source. The domain is unbounded, thus we append perfectly matched layers \cite{bermudez2007optimal} to the computation domain. We use the default GMRES method to solve \eqref{eq:helmholtz_eq}. 

\subsection{Measurement Setup}

\begin{figure}
    \centering
    \includegraphics[width=1\linewidth]{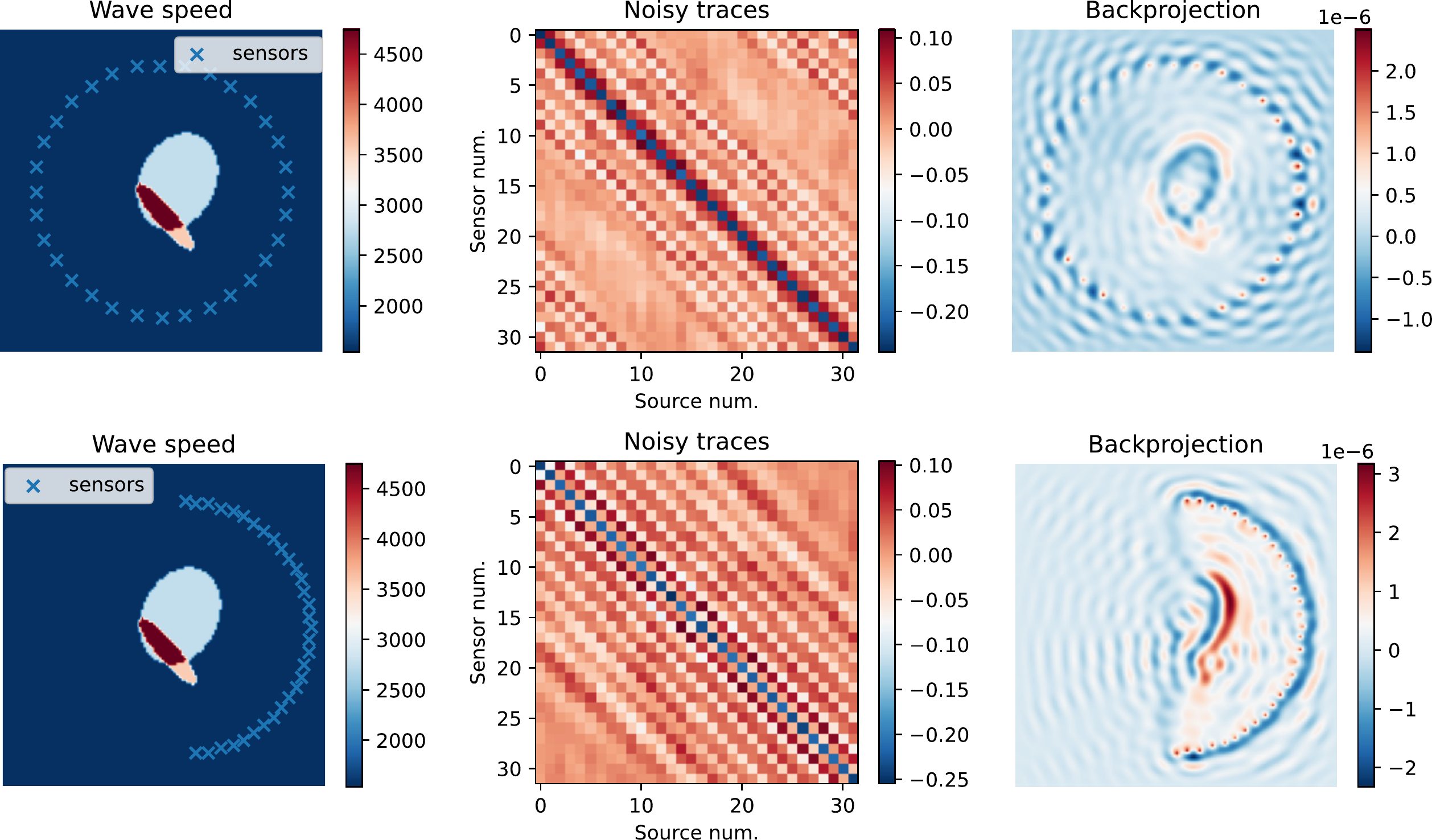}
    \caption{Examples of the inverse medium scattering problem. Top row: full-view measurement. Bottom row: limited-view measurement. We add $30~dB$ noise to the measurements. The backprojections are computed from auto-differentiation.}
    \label{fig:measurement_setup}
\end{figure}

To image the heterogeneity, we place the (collocated) sources and sensors on a circle. Each source takes a turn to emit a circular wave that scatters through the medium and gets measured at all the sensors. Thus our measurement data are square matrices of complex-valued entries. 

The inverse problem is to recover the medium from the measurement data. To set up the notation, let us denote the unknown parameters as ${\rvx:=c|_{\Omega}}$ and the (backprojected) measurements as ${\rvy:=u|_{\partial \Omega}}$. Fig.~\ref{fig:measurement_setup} illustrates two examples of the measurement setup.

The \verb+j-wave+ package \cite{stanziola2022jwave} gives easy access to the backprojection (BP) images thanks to automatic differentiation. The BP is obtained by applying the Jacobian of the discrete forward model to the measurement mismatch. This auto-diff is the main component in the discretize-then-optimize regime. The advantage is that the Jacobian exactly matches the numerical model being used and it can easily extend to other medium parameters such as density and attenuation.

The BP image is computed by taking the derivative of the measurement loss
$$
    \rvy \mapsto \dfrac{\partial }{\partial x} \|\rvy - \hat{y}(x)\|^2_2.
$$
With a slight abuse of notation, we will use these BP images, denoted as $\rvy$, to estimate the posterior distribution of medium wave speed.

\subsection{Training Data}

We generate 4000 medium samples, each containing a set of random ellipsoidal scatterers, along with their discrete boundary measurements. The computation domain is a $128\times 128$ grid with resolution $\Delta x = 10^{-3}~\mathrm{m}$. The background wave speed is $1540~\mathrm{m/s}$ and the maximum contrast is close to $4$ times the background. The angular frequency is ${\omega=7\cdot 10^{5}~\mathrm{s}^{-1}}$, which corresponds to $13$ grid points per wavelength. We use 32 complex-valued measurements corrupted with Gaussian noise.

\section{Amortized Inference for Inverse Problems}

Variational inference~\cite{jordan1999introduction, rezende2015variational,papamakarios2021normalizing} is a technique for approximating posterior distributions $p_{X|Y}(x)$ using a parametrized family of normalized densities $q_\theta (x)$. To determine $\theta$, we minimize the \textit{forward} Kullback–Leibler (KL) divergence~\cite{papamakarios2021normalizing,bishop2006pattern},
\begin{equation}
    \theta^\ast(y) = \argmin_{\theta \in \Theta} ~ \text{KL}(p_{X|Y}( \cdot \, | \, y) \| q_\theta ).
\end{equation}
This expression implies that for each measurement $\rvy$, we have to solve an optimization problem to obtain $\theta^\ast(\rvy)$. We are interested in measurements for many samples $\rvx$, which would require many minimizations of the above expression. Thus, we choose to \textit{amortize} inference by defining a bivariate family $q_\theta(x,y)$. Amortized inference minimizes the KL divergence averaged over all samples simultaneously~\cite{ whang2021composing, khorashadizadeh2022conditional},
\begin{equation} \label{eq:amortized inference}
\begin{aligned}
    \theta^*
    &= \argmin_{\theta} ~  \mathbb{E}_{Y \sim p_Y}  \text{KL}(p_{X|Y}( \, \cdot\, |Y) \| q_\theta(\, \cdot\, |Y)) \\
    &= \argmax_{\theta} ~ \mathbb{E}_{X, Y \sim p_{X, Y}} \log q_{\theta}(X|Y).
\end{aligned}
\end{equation}
The expectation over $P_{X|Y}(x|y)$, the density in question, is estimated by the empirical expectation over training data $\{(\rvx_i,\rvy_i)\}_{i = 1}^N$.
In this paper, the variational approximator $q_\theta$ is a neural network.

\section{U-Flow}
\label{sec:U-Flow}
\begin{figure*}
    \centering
    \includegraphics[width=0.8\linewidth]{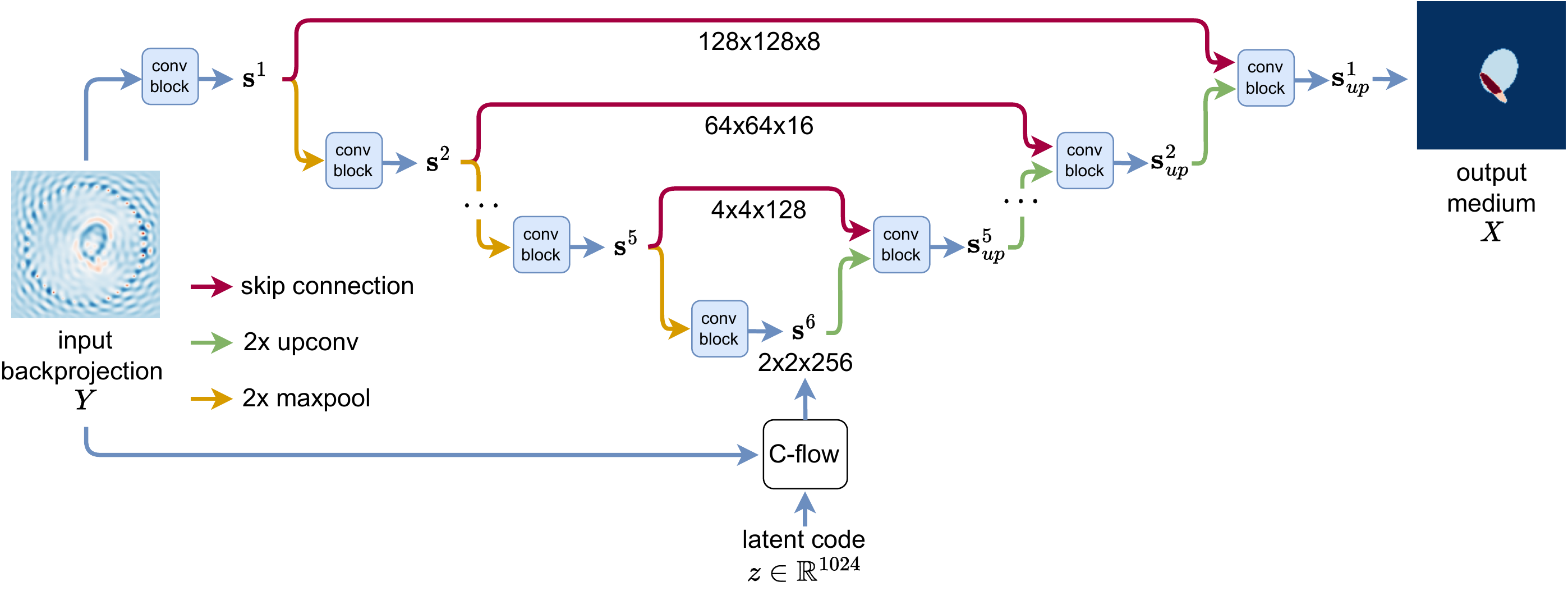}
    \caption{Network architecture of U-Flow. In our implementation, the U-Net component has 6 scale levels $(\rvs^1,\rvs^2, \dots, \rvs^6)$. At the coarsest level, the latent variable $\rvs^6$ is a ${2\times2\times256}$ tensor, which is the output of a conditional flow model, conditioned by original input.}
    \label{fig:uflow_diagram}
\end{figure*}

We propose U-Flow, a probabilistic model combining
U-Net~\cite{ronneberger2015u} and conditional normalizing flows~\cite{ardizzone2019guided} to approximate the posterior distribution.

\subsection{U-Net}
\label{sec:Unet}
Ronneberger \emph{et al.} originally developed the U-Net for medical image segmentation. It has since been adapted to many image-to-image tasks, often achieving (near-)state-of-the-art performance. The U-Net is an encoder-decoder network
$$\text{UNet}_\phi \eqdef \text{dec} \circ \text{enc}.$$
The encoder and the decoder are both convolutional neural networks with pooling layers. The encoder takes the measurements $\rvy$ as input and produces features in different scales. The decoder then takes the computed features and reconstructs the target signal $\rvx$.

The encoder and decoder are jointly optimized using the mean-square error (MSE) loss
\begin{equation}
    \phi^* = \argmin_{\phi} ~ \dfrac{1}{N} \sum_{i = 1}^N \| \rvx_i - \text{UNet}_{\phi}(\rvy_i) \|_2^2.
    \label{eq: unet_loss}
\end{equation}

While the U-Net produces high-quality reconstructions, its output is a single estimate. In this paper, we present a probabilistic version of the U-Net for amortized Bayesian inference.

\subsection{Conditional Normalizing Flows}
\label{sec:c-flow}
Normalizing flows~\cite{dinh2016density,kingma2018glow,kothari2021trumpets} are a class of likelihood-based generative models. They transform a simple and known distribution into an unknown data distribution by a sequence of invertible mappings. Invertibility enables efficient likelihood estimation and maximum-likelihood (ML) parameter fitting. Vlašić et al. \cite{vlasic2022implicitObstacleScattering} demonstrated the effectiveness of normalizing flow-based generative models in regularizing the inverse obstacle scattering problem. Recent works proposed conditional versions of normalizing flows~\cite{ardizzone2019guided} that allow for approximation of posterior distributions. However, regular conditional flows require the latent space dimension to equal the data dimension, which leads to a large network and slow training. Moreover, as the range of conditional flows covers the entire space, the posterior samples are not constrained to an image distribution and are often of low quality in ill-posed nonlinear inverse problems~\cite{khorashadizadeh2022conditional}.

\subsection{Our Approach}
\label{sec:our_approach}
Since the input and output of flow models must have the same dimension, it is opportune to  use flows to model low-dimensional latent spaces rather than images directly. We use flows to approximate the posterior distribution of the coarsest scale in the U-Net. Concretely, as shown in Fig.~\ref{fig:uflow_diagram}, the encoder of the U-Net takes the BPs and produces features in six scales,  $\text{enc}(\rvy) = (\rvs^1,\rvs^2, \dots, \rvs^6)$ for ${\rvy \in \mathbb{R}^{D}}$. These multiscale features feed to the decoder to reconstruct a \textit{single} estimate of the target signal ${\hat{\rvx}(\rvy) = \text{dec}(\rvs^6, \rvs^5, \dots, \rvs^1)}$. To generate posterior samples, we let a flow model learn the conditional distribution of features at the coarsest scale $p_{\rvs^6|Y}$ where ${\rvs^6 \in \R^d}$ and ${d \ll D}$. We first train a U-Net with the loss in \eqref{eq: unet_loss} and compute the coarsest scale features of the BPs samples in the training set. Having obtained a paired training set of the BPs and the corresponding features $\{(\rvs^6_i, \rvy_i)\}_{i = 1}^N$, we then train a flow model. We use the conditional version of Glow~\cite{kingma2018glow}, where we deploy conditional coupling blocks proposed in~\cite{ardizzone2019guided} to condition the generation on backprojections. The conditional flow model is trained using amortized inference loss~\eqref{eq:amortized inference} as,
\begin{equation}
    \theta^* = \argmin_{\theta} \dfrac{1}{N}\sum_{i=1}^N \left(-\log p_Z(\rvz_i) + \log|\det J_{f_{\theta}}|\right),
    \label{eq: U-Flow loss}
\end{equation}
where ${\rvz_i = f_{\theta}^{-1}(\rvs_i^6 , \rvy_i)}$, $J_{f_{\theta}}$ is the Jacobian matrix of $f_\theta$ and $p_Z$ is a multivariate Gaussian distribution.
In this paper, instead of directly approximating $p_{X|Y}$, we approximate the posterior distribution of the features in the lowest scale of the U-Net $p_{s^6_i|Y}$ using conditional normalizing flows as shown in Fig.~\ref{fig:uflow_diagram}.
When the conditional flow model is trained, we can generate posterior samples for each BP $y^*$,
\begin{equation}
    \rvx_{\text{post}} (\rvy^*) = \text{dec}(f_\theta(\rvz), \rvs^5, ..., \rvs^1)
    \label{eq: U-Flow posterior}
\end{equation}
where ${\rvz \sim \mathcal{N}(0,I)}$ and ${(\rvs^1, ..., \rvs^5, \cdot) = {\text{enc}(\rvy^*)}}$.
The key advantage of the proposed model is that the posterior samples have a low-dimensional structure, which acts as a strong regularizer for ill-posed inverse problems.

\section{Results}
\begin{figure*}
\centering
\subfloat[Limited-view: the receivers and incident waves are only on the right-hand side of the object; U-Flow could reliably capture a physically meaningful uncertainty estimate by showing more uncertainty (red regions) on the left part of the object.]{\includegraphics[width=1.3\columnwidth]{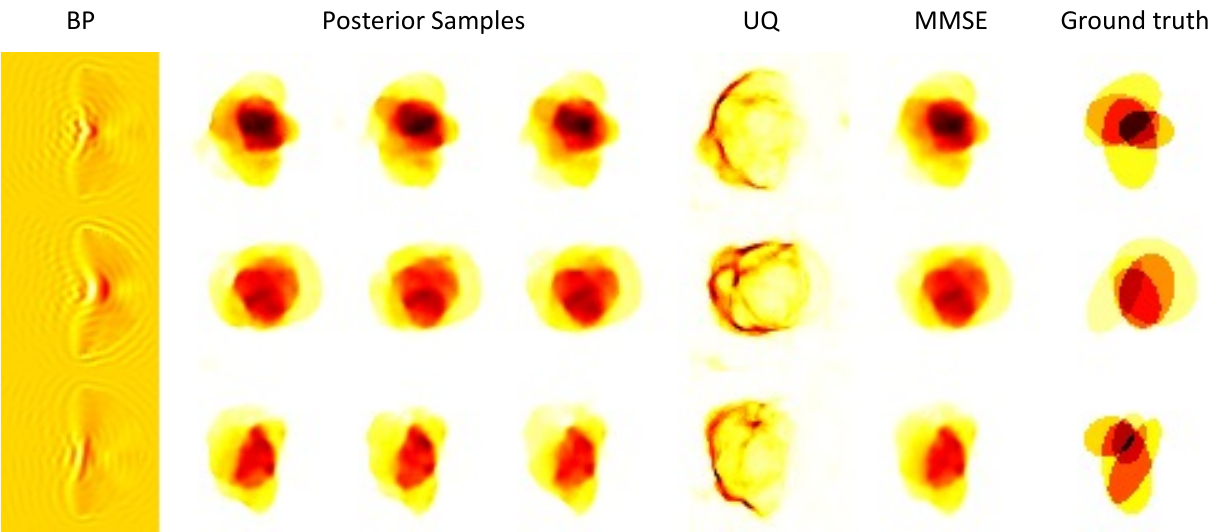}%
\label{fig:side-view}}
\vfil
\subfloat[Full-view: the receivers and incident waves are uniformly distributed around the object. ]{\includegraphics[width=1.3\columnwidth]{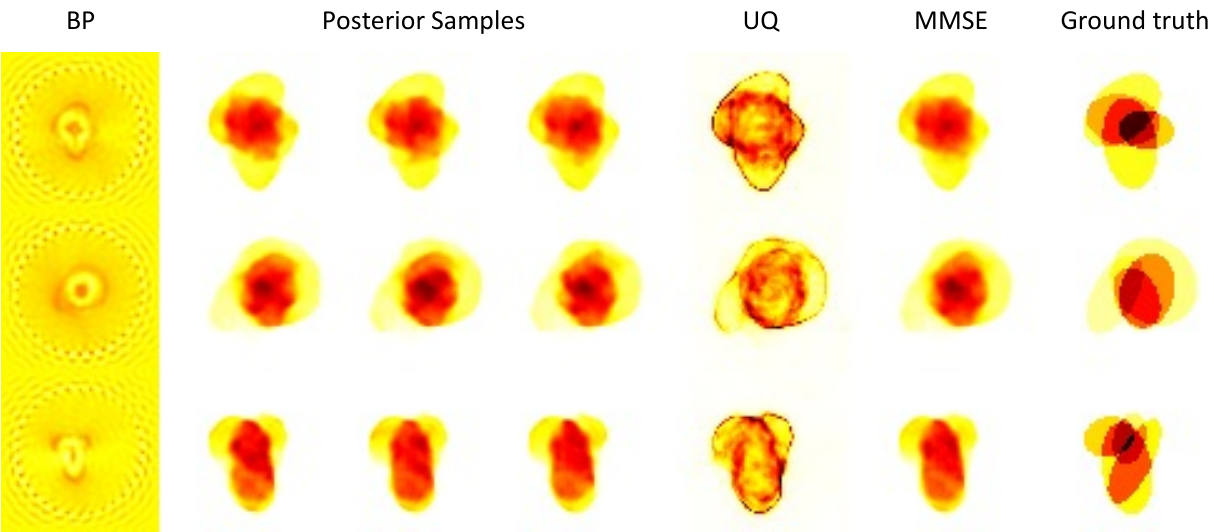}%
\label{fig:full-view}}
\caption{Performance of U-Flow over inverse medium scattering}
\label{fig:main results}
\end{figure*}

\begin{figure}
    \centering
    \includegraphics[width=1\linewidth]{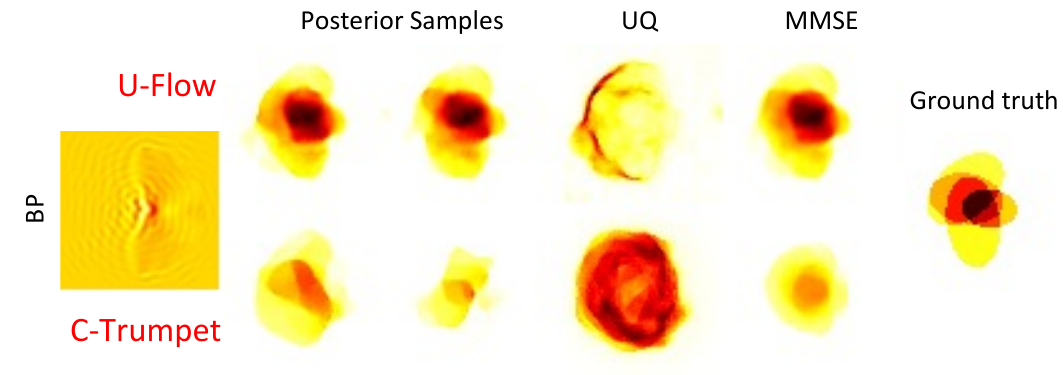}
    \caption{Performance comparison of U-Flow and C-Trumpet~\cite{khorashadizadeh2022conditional} for the limited-view problem (no sensors on the left-hand side); U-Flow outperforms C-Trumpet in both posterior sampling and uncertainty quantification by assigning more uncertainty in the left part of the object (red regions).}
    \label{fig:performance comparison}
\end{figure}

We trained U-Flow for 600 epochs in total, 300 for the U-Net and 300 for the conditional flow model. We used the Adam optimizer~\cite{kingma2014adam} with the learning rate set to $10^{-4}$. The conditional flow model was composed of $24$ Glow-blocks, each containing activation normalization, the $1 \times 1$ convolution, and a conditional coupling layer. We compared U-Flow with C-Trumpet~\cite{khorashadizadeh2022conditional}, a conditional injective flow which 
is well-suited for solving ill-posed inverse problems. In our experiments, U-Flow and C-Trumpet had $9$M and $14$M trainable parameters, respectively.
The MMSE estimate was calculated by averaging $25$ posterior samples. The UQ is performed by dividing the pixel-wise standard deviation on $25$ posterior samples to the MMSE estimate to show the relative error. 

Fig.~\ref{fig:side-view} illustrates the performance of U-Flow on inverse medium scattering with a limited-view sensing configuration, where the receivers and incident waves are located on the \textit{right-hand} side of the observed medium. This experiment shows that U-Flow can generate various posterior samples and capture a physically meaningful UQ. Notice that U-Flow assigned more uncertainty to the \textit{left-hand} side of the medium (red region).
The results for the full-view configuration are shown in Fig.~\ref{fig:full-view}. As expected, the full-view configuration yields better posterior samples than the limited-view configuration.

Fig.~\ref{fig:performance comparison} compares the performance of U-Flow and C-Trumpet. The experiment shows that U-Flow significantly outperforms C-Trumpet both in posterior sampling and UQ. Table~\ref{tab:quantitative results} gives a quantitative comparison of U-Flow with baselines, including the basic U-Net~\cite{ronneberger2015u}. U-Flow exhibits comparable results to the U-Net while giving access to posterior samples and UQ.

\begin{table}
\renewcommand{\arraystretch}{1.3}
\caption{SNR of MMSE estimate (computed over 25 posterior samples for flow-based models) of different models over inverse medium scattering in two setups}
\label{tab:quantitative results}
\centering
\resizebox{0.45\textwidth}{!}{%
\begin{tabular}{@{}lcccc@{}}
\hline
& BP & C-Trumpets~\cite{khorashadizadeh2022conditional} & U-Net~\cite{ronneberger2015u} & U-Flow \\
\hline
{\textit{Full-view}}   & -17.8 & 5.7  & \textbf{11.3} & \textbf{11.3}  \\
{\textit{Side-view}} & -17.7 &  5.2  & \textbf{9.7}  & 9.4 \\
\hline
\end{tabular}}
\end{table}

\section{Conclusion}

We demonstrated that the dichotomy between high-quality inversions without UQ by standard point-estimate networks, and low-quality inversions with UQ by the various conditional generative models is a false one. By combining a low-dimensional flow with a U-Net, we get the best of both worlds. The reconstructions are very fast---orders of magnitude faster than with the standard iterative methods. It will be interesting to see how the proposed model performs on other inverse problems.

\section*{Acknowledgment}
A.~K., H.~N., and I.~D. acknowledge support from the European Research Council under Starting Grant 852821--SWING.

\bibliographystyle{IEEEtran}
\bibliography{main}

\begin{thebibliography}{10}
\providecommand{\url}[1]{#1}
\csname url@samestyle\endcsname
\providecommand{\newblock}{\relax}
\providecommand{\bibinfo}[2]{#2}
\providecommand{\BIBentrySTDinterwordspacing}{\spaceskip=0pt\relax}
\providecommand{\BIBentryALTinterwordstretchfactor}{4}
\providecommand{\BIBentryALTinterwordspacing}{\spaceskip=\fontdimen2\font plus
\BIBentryALTinterwordstretchfactor\fontdimen3\font minus
  \fontdimen4\font\relax}
\providecommand{\BIBforeignlanguage}[2]{{%
\expandafter\ifx\csname l@#1\endcsname\relax
\typeout{** WARNING: IEEEtran.bst: No hyphenation pattern has been}%
\typeout{** loaded for the language `#1'. Using the pattern for}%
\typeout{** the default language instead.}%
\else
\language=\csname l@#1\endcsname
\fi
#2}}
\providecommand{\BIBdecl}{\relax}
\BIBdecl

\bibitem{cheney2009fundamentals}
M.~Cheney and B.~Borden, \emph{Fundamentals of radar imaging}.\hskip 1em plus
  0.5em minus 0.4em\relax SIAM, 2009.

\bibitem{amin2017through}
M.~G. Amin, \emph{Through-the-wall radar imaging}.\hskip 1em plus 0.5em minus
  0.4em\relax CRC press, 2017.

\bibitem{persico2014introduction}
R.~Persico, \emph{Introduction to ground penetrating radar: inverse scattering
  and data processing}.\hskip 1em plus 0.5em minus 0.4em\relax John Wiley \&
  Sons, 2014.

\bibitem{zoughi2000microwave}
R.~Zoughi, \emph{Microwave non-destructive testing and evaluation
  principles}.\hskip 1em plus 0.5em minus 0.4em\relax Springer Science \&
  Business Media, 2000, vol.~4.

\bibitem{chen2018computational}
X.~Chen, \emph{Computational methods for electromagnetic inverse
  scattering}.\hskip 1em plus 0.5em minus 0.4em\relax John Wiley \& Sons, 2018.

\bibitem{nachman1996global}
A.~I. Nachman, ``Global uniqueness for a two-dimensional inverse boundary value
  problem,'' \emph{Annals of Mathematics}, pp. 71--96, 1996.

\bibitem{sun2021deep}
H.~Sun and K.~L. Bouman, ``Deep probabilistic imaging: Uncertainty
  quantification and multi-modal solution characterization for computational
  imaging,'' in \emph{Proceedings of the AAAI Conference on Artificial
  Intelligence}, vol.~35, no.~3, 2021, pp. 2628--2637.

\bibitem{khorashadizadeh2022conditional}
A.~Khorashadizadeh, K.~Kothari, L.~Salsi, A.~A. Harandi, M.~de~Hoop, and
  I.~Dokmanić, ``Conditional injective flows for {Bayesian} imaging,''
  \emph{arXiv:2204.07664}, 2022.

\bibitem{begoli2019need}
E.~Begoli, T.~Bhattacharya, and D.~Kusnezov, ``The need for uncertainty
  quantification in machine-assisted medical decision making,'' \emph{Nature
  Machine Intelligence}, vol.~1, no.~1, pp. 20--23, 2019.

\bibitem{stoyer2009science}
M.~Stoyer, D.~McNabb, J.~Burke, and L.~Bernstein, ``Science based stockpile
  stewardship, uncertainty quantification, and surrogate reactions,'' Lawrence
  Livermore National Lab.(LLNL), Livermore, CA (United States), Tech. Rep.,
  2009.

\bibitem{brown2015uncertainty}
D.~Brown, M.~Herman, S.~Hoblit, E.~McCutchan, G.~Nobre, B.~Pritychenko, and
  A.~Sonzogni, ``Uncertainty quantification in the nuclear data program,''
  \emph{Journal of Physics G: nuclear and particle physics}, vol.~42, no.~3, p.
  034020, 2015.

\bibitem{arnez2020comparison}
F.~Arnez, H.~Espinoza, A.~Radermacher, and F.~Terrier, ``A comparison of
  uncertainty estimation approaches in deep learning components for autonomous
  vehicle applications,'' \emph{arXiv preprint arXiv:2006.15172}, 2020.

\bibitem{michelmore2018evaluating}
R.~Michelmore, M.~Kwiatkowska, and Y.~Gal, ``Evaluating uncertainty
  quantification in end-to-end autonomous driving control,'' \emph{arXiv
  preprint arXiv:1811.06817}, 2018.

\bibitem{tarantola2005inverse}
A.~Tarantola, \emph{Inverse problem theory and methods for model parameter
  estimation}.\hskip 1em plus 0.5em minus 0.4em\relax SIAM, 2005.

\bibitem{stuart2010inverse}
A.~M. Stuart, ``Inverse problems: a {Bayesian} perspective,'' \emph{Acta
  numerica}, vol.~19, pp. 451--559, 2010.

\bibitem{martin2012stochastic}
J.~Martin, L.~C. Wilcox, C.~Burstedde, and O.~Ghattas, ``A stochastic newton
  mcmc method for large-scale statistical inverse problems with application to
  seismic inversion,'' \emph{SIAM Journal on Scientific Computing}, vol.~34,
  no.~3, pp. A1460--A1487, 2012.

\bibitem{zhao2019gradient}
Z.~Zhao and M.~K. Sen, ``A gradient based mcmc method for fwi and uncertainty
  analysis,'' in \emph{SEG International Exposition and Annual Meeting}.\hskip
  1em plus 0.5em minus 0.4em\relax OnePetro, 2019.

\bibitem{cui2015data}
T.~Cui, Y.~M. Marzouk, and K.~E. Willcox, ``Data-driven model reduction for the
  {Bayesian} solution of inverse problems,'' \emph{International Journal for
  Numerical Methods in Engineering}, vol. 102, no.~5, pp. 966--990, 2015.

\bibitem{peherstorfer2018survey}
B.~Peherstorfer, K.~Willcox, and M.~Gunzburger, ``Survey of multifidelity
  methods in uncertainty propagation, inference, and optimization,'' \emph{Siam
  Review}, vol.~60, no.~3, pp. 550--591, 2018.

\bibitem{esser2018variational}
P.~Esser, E.~Sutter, and B.~Ommer, ``A variational u-net for conditional
  appearance and shape generation,'' in \emph{Proceedings of the IEEE
  conference on computer vision and pattern recognition}, 2018, pp. 8857--8866.

\bibitem{jin2019fast}
L.~Jin, H.~Lu, and G.~Wen, ``Fast uncertainty quantification of reservoir
  simulation with variational u-net,'' \emph{arXiv preprint arXiv:1907.00718},
  2019.

\bibitem{kingma2013auto}
D.~P. Kingma and M.~Welling, ``Auto-encoding variational bayes,'' \emph{arXiv
  preprint arXiv:1312.6114}, 2013.

\bibitem{siahkoohi2022deep}
A.~Siahkoohi, G.~Rizzuti, and F.~J. Herrmann, ``Deep {Bayesian} inference for
  seismic imaging with tasks,'' \emph{Geophysics}, vol.~87, no.~5, pp.
  S281--S302, 2022.

\bibitem{wei2020uncertainty}
Z.~Wei and X.~Chen, ``Uncertainty quantification in inverse scattering problems
  with {Bayesian} convolutional neural networks,'' \emph{IEEE Transactions on
  Antennas and Propagation}, vol.~69, no.~6, pp. 3409--3418, 2020.

\bibitem{ardizzone2019guided}
L.~Ardizzone, C.~L{\"u}th, J.~Kruse, C.~Rother, and U.~K{\"o}the, ``Guided
  image generation with conditional invertible neural networks,'' \emph{arXiv
  preprint arXiv:1907.02392}, 2019.

\bibitem{siahkoohi2020faster}
A.~Siahkoohi, G.~Rizzuti, P.~A. Witte, and F.~J. Herrmann, ``Faster uncertainty
  quantification for inverse problems with conditional normalizing flows,''
  \emph{arXiv preprint arXiv:2007.07985}, 2020.

\bibitem{zhao2022bayesian}
X.~Zhao, A.~Curtis, and X.~Zhang, ``{Bayesian} seismic tomography using
  normalizing flows,'' \emph{Geophysical Journal International}, vol. 228,
  no.~1, pp. 213--239, 2022.

\bibitem{ronneberger2015u}
O.~Ronneberger, P.~Fischer, and T.~Brox, ``U-net: Convolutional networks for
  biomedical image segmentation,'' in \emph{International Conference on Medical
  image computing and computer-assisted intervention}.\hskip 1em plus 0.5em
  minus 0.4em\relax Springer, 2015, pp. 234--241.

\bibitem{stanziola2022jwave}
A.~Stanziola, S.~R. Arridge, B.~T. Cox, and B.~E. Treeby, ``j-wave: An
  open-source differentiable wave simulator,'' \emph{arXiv preprint
  arXiv:2207.01499}, 2022.

\bibitem{bermudez2007optimal}
A.~Berm{\'u}dez, L.~Hervella-Nieto, A.~Prieto, R.~Rodr{\i} \emph{et~al.}, ``An
  optimal perfectly matched layer with unbounded absorbing function for
  time-harmonic acoustic scattering problems,'' \emph{Journal of computational
  Physics}, vol. 223, no.~2, pp. 469--488, 2007.

\bibitem{jordan1999introduction}
M.~I. Jordan, Z.~Ghahramani, T.~S. Jaakkola, and L.~K. Saul, ``An introduction
  to variational methods for graphical models,'' \emph{Machine learning},
  vol.~37, no.~2, pp. 183--233, 1999.

\bibitem{rezende2015variational}
D.~Rezende and S.~Mohamed, ``Variational inference with normalizing flows,'' in
  \emph{International conference on machine learning}.\hskip 1em plus 0.5em
  minus 0.4em\relax PMLR, 2015, pp. 1530--1538.

\bibitem{papamakarios2021normalizing}
G.~Papamakarios, E.~T. Nalisnick, D.~J. Rezende, S.~Mohamed, and
  B.~Lakshminarayanan, ``Normalizing flows for probabilistic modeling and
  inference.'' \emph{J. Mach. Learn. Res.}, vol.~22, no.~57, pp. 1--64, 2021.

\bibitem{bishop2006pattern}
C.~M. Bishop and N.~M. Nasrabadi, \emph{Pattern recognition and machine
  learning}.\hskip 1em plus 0.5em minus 0.4em\relax Springer, 2006, vol.~4,
  no.~4.

\bibitem{whang2021composing}
J.~Whang, E.~Lindgren, and A.~Dimakis, ``Composing normalizing flows for
  inverse problems,'' in \emph{International Conference on Machine
  Learning}.\hskip 1em plus 0.5em minus 0.4em\relax PMLR, 2021, pp.
  11\,158--11\,169.

\bibitem{dinh2016density}
L.~Dinh, J.~Sohl-Dickstein, and S.~Bengio, ``Density estimation using real
  nvp,'' \emph{arXiv preprint arXiv:1605.08803}, 2016.

\bibitem{kingma2018glow}
D.~P. Kingma and P.~Dhariwal, ``Glow: Generative flow with invertible 1x1
  convolutions,'' \emph{Advances in neural information processing systems},
  vol.~31, 2018.

\bibitem{kothari2021trumpets}
K.~Kothari, A.~Khorashadizadeh, M.~de~Hoop, and I.~Dokmani{\'c}, ``Trumpets:
  Injective flows for inference and inverse problems,'' in \emph{Uncertainty in
  Artificial Intelligence}.\hskip 1em plus 0.5em minus 0.4em\relax PMLR, 2021,
  pp. 1269--1278.

\bibitem{vlasic2022implicitObstacleScattering}
T.~{Vla{\v{s}}i{\'c}}, H.~{Nguyen}, A.~{Khorashadizadeh}, and
  I.~{Dokmani{\'c}}, ``{Implicit Neural Representation for Mesh-Free Inverse
  Obstacle Scattering},'' \emph{arXiv:2206.02027}, 2022.

\bibitem{kingma2014adam}
D.~P. Kingma and J.~Ba, ``Adam: A method for stochastic optimization,''
  \emph{arXiv preprint arXiv:1412.6980}, 2014.

\end{thebibliography}

\end{document}